\documentclass[envcountsame]{llncs}
\usepackage{amsmath}
\usepackage{amssymb}
\usepackage{amsfonts}
\usepackage[dvips,ps,all]{xy}
\usepackage{graphicx}
\usepackage{subfig}

\newenvironment{Theorem}{\begin{theorem}}{\end{theorem}}

\newenvironment{Lemma}{\begin{lemma}}{\end{lemma}}

\newenvironment{Remark}{\begin{remark}\rm}{\qed\end{remark}}

\newenvironment{Example}{\begin{example}\rm}{\end{example}}
\newenvironment{Definition}{\begin{definition}\rm}{\end{definition}}
\newenvironment{Proof}{\begin{proof}}{\qed\end{proof}}

\begin{document}

\newcommand{\remconf}[1]{#1}
\newcommand{\addconf}[1]{}

\remconf{
\title{Non-Confluent NLC Graph Grammar Inference by Compressing Disjoint Subgraphs}}
\addconf{
\title{Learning Non-Confluent NLC Graph Grammar Rules}}
\author{Hendrik Blockeel\inst{1,2} \and Robert Brijder\inst{1}}

\institute{Leiden Institute of Advanced Computer Science,
Universiteit Leiden,\\ The Netherlands, \email{rbrijder@liacs.nl}
\and Department of Computer Science, Katholieke Universiteit
Leuven,\\ Belgium, \email{hendrik.blockeel@cs.kuleuven.be}}

\bibliographystyle{plain}
\maketitle

\newcommand{\spair}{(S_1,S_2)}
\newcommand{\spairinv}{(S_2,S_1)}
\newcommand{\incomplete}[1]{}

\begin{abstract}
Grammar inference deals with determining (preferable simple)
models/grammars consistent with a set of observations. There is a
large body of research on grammar inference within the theory of
formal languages. However, there is surprisingly little known on
grammar inference for graph grammars. In this paper we take a
further step in this direction and work within the framework of node
label controlled (NLC) graph grammars. Specifically, we
characterize, given a set of disjoint and isomorphic subgraphs of a
graph $G$, whether or not there is a NLC graph grammar rule which
can generate these subgraphs to obtain $G$. This generalizes
previous results by assuming that the set of isomorphic subgraphs is
disjoint instead of non-touching. This leads naturally to consider
the more involved ``non-confluent'' graph grammar rules.
\end{abstract}

\section{Introduction}
Grammar inference, also called grammar induction, is a general line
of research where one is concerned with determining a ``simple''
grammar that is consistent with a given set of possible and
impossible outcomes. Hence, one ``goes back'' in the derivation:
instead of determining the generative power of a grammar, one
determines the grammar given the generated output. This topic is
well-studied for formal languages, especially with respect to
context-free languages, see e.g. \cite{Vidal_1994,Fu_Booth_1_1986},
however, relatively little is known for graph grammars.

The topic of inference of graph grammars is considered in
\cite{Jonyer_Holder_Cook_2004} and uses their so-called Subdue
scheme developed in \cite{Cook_Holder_1994}. In
\cite{Blockeel_NLC_Induction} a rigorous approach of grammar
inference within the framework of node label controlled (NLC) graph
grammars \cite{Engelfriet_Introduction_NLC}, a natural and
well-studied class of graph grammars, is initiated. There it is
characterized, given a set $\mathcal{S}$ of non-touching isomorphic
graphs of a graph $G$, whether or not there is a graph grammar
consisting of one rule able to generate the graphs of $\mathcal{S}$
to obtain $G$. We continue this research and generalize this result
for the case where these graphs are disjoint instead of
non-touching. Such a generalization requires one to deal with a
number of issues. Most notably, one has to deal with non-confluency
issues: the generated graph depends on the order in which touching
subgraphs are generated.
%

\addconf{Due to space constraints, proofs of the results are
omitted, but can be found in an extended version .. of this paper.}

\section{Notation and Terminology}
We consider (simple) graphs $G = (V,E)$, where $V$ is a finite set
of nodes and $E \subseteq \{\{x,y\} \mid x,y \in V, x \not= y\}$ is
the set of edges -- hence no loops or parallel edges are allowed. We
denote $V(G) = V$ and $E(G) = E$. For $S \subseteq V$, the induced
subgraph of $G$ is $(S,E')$ where $E' \subseteq E$ and for each $e
\in E$ we have $e \in E'$ iff $e \subseteq S$. We consider only
induced subgraphs, and therefore we sometimes just write
``subgraph'' instead of induced subgraph. The neighborhood of $S
\subseteq V$ in $G$, denoted by $N_G(S)$, is $\{ v \in V \backslash
S \mid \{s,v\} \in E \mbox{ for some } s \in S\}$. If $S = \{x\}$ is
a singleton, then we also write $N_G(x) = N_G(S)$. A labelled graph
is a triple $G = (V,E,l)$ where $(V,E)$ is a graph and $l : V
\rightarrow L$ is a node labelling function, where $L$ is a finite
set of labels. As usual, graphs are consider isomorphic if they are
identical modulo the identity of the vertex. It is important to
realize that for labelled graphs, vertices identified by an
isomorphism have identical labels. In graphical depictions of
labelled graphs we will always represent the vertices by their
labels.

Subgraphs $G_1$, and $G_2$ are called \emph{disjoint} if $V(G_1)$
and $V(G_2)$ are disjoint. They are called \emph{touching} if
$V(G_1) \cup N_G(V(G_1))$ and $V(G_2) \cup N_G(V(G_2))$ are not
disjoint.

Define, for disjoint $W_1, W_2 \subseteq V$, $K_{W_1,W_2} = \{
(x_1,x_2) \mid x_1 \in W_1, x_2 \in W_2\}$ to be the set of all
tuples with the left element from $W_1$ and the right element from
$W_2$. Define $u((x_1,x_2))$ to be the underlying set $\{x_1,x_2\}$,
and define $\pi_i((x_1,x_2)) = x_i$ for $i \in \{1,2\}$. Often, for
function $f : X \rightarrow Y$ we write $f(D) = \{ f(x) \mid x \in
D\}$ for $D \subseteq X$.

\section{NLC Graph Grammars}
Typically, a graph grammar transforms a graph $G$ by replacing an
(induced) subgraph $H$ by another graph $H'$ where $H'$ is embedded
in the remaining part $G \backslash H$ of the original graph in a
way prescribed by a so-called graph grammar embedding relation. The
node label controlled (NLC) graph grammars are the simplest class of
these grammars, where $H$ is a single node. Note that for the
grammars the exact identities of the nodes are not important as
multiple copies of $H'$ may be inserted. Hence, we consider labelled
graphs where the embedding relation is defined w.r.t. node labels
instead of nodes. In this section we recall informally the notions
and definitions concerning NLC grammars used in this paper, and
refer to \cite{Engelfriet_Introduction_NLC} for a gentle and more
detailed introduction to these grammars.

A NLC graph grammar is a system $Q$ consisting of a set of node
labels $L$, an \emph{embedding relation} $E \subseteq L^2$, and a
set of \emph{productions} $P$ where a production is of the form $N
\rightarrow S$ where $N \in L$ and $S$ is a (labelled) graph. In
this paper
\incomplete{, with the exception of
Section~\ref{sec_multiple_prod},}
we will focus on the case $|P| = 1$. Hence $Q$ can be denoted as a
\emph{rule} $r = N \rightarrow S / E$ (if $L$ is understood from the
context of considerations). Given a graph $G$, $r$ can be applied to
any node $v$ labelled by $N$. The result of applying $r$ to $v$ in
$G$ is that $v$ is removed from $G$ along with the edges adjacent to
$v$, and (a copy of) $S$ is added to $G$, and an edge $e = \{x,y\}$
is added to $G$ iff $x \in V(S)$, $y \in N_G(S)$ and $(l(x),l(y))
\in E$ (recall that $l$ is the labelling function). To avoid
confusion with embedding relations, the set of edges of a graph $G$
are written in the remainder as $E(G)$ and not as $E$.

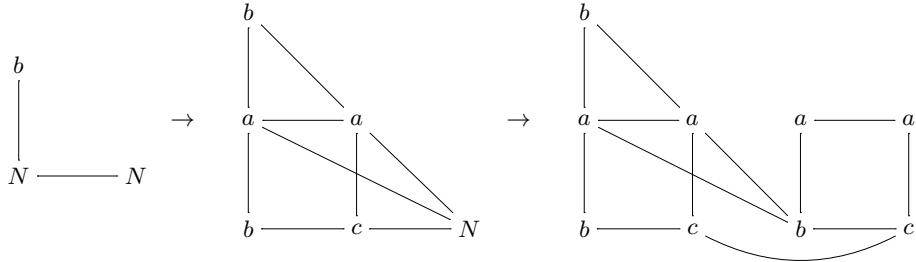
\begin{figure}
\begin{minipage}[c]{50pt} $\xymatrix@=30pt{
b \ar@{-}[d] & \\
N \ar@{-}[r] & N
}$\end{minipage}
\hspace{0.3cm} $\rightarrow$  \hspace{0.3cm}
%
\begin{minipage}[c]{100pt}
$ \xymatrix@=30pt{
b \ar@{-}[rd] \ar@{-}[d] & \\
a \ar@{-}[r] \ar@{-}[drr] & a \ar@{-}[d] \ar@{-}[dr] & \\
b \ar@{-}[u] & c \ar@{-}[l] \ar@{-}[r] & N
}$
\end{minipage}
%
$\rightarrow$ \hspace{0.3cm}
\begin{minipage}[c]{100pt}
$ \xymatrix @=30pt{
b \ar@{-}[rd] \ar@{-}[d] & \\
a \ar@{-}[r] \ar@{-}[drr] & a \ar@{-}[d] \ar@{-}[dr] & a \ar@{-}[r] & a \ar@{-}[d] \\
b \ar@{-}[u] & c \ar@{-}[l] \ar@{-}@/_1.0pc/[rr] & b \ar@{-}[u] & c
\ar@{-}[l]
} $ \end{minipage} \caption{The derivation of a graph $G$ (left-hand
side) to $G'$ (right-hand side).} \label{ex_fig_touching}
\end{figure}

\begin{Example} \label{ex_nlc_grammar}
Let $G$ be the graph on the left-hand side of
Figure~\ref{ex_fig_touching}. Consider the grammar rule $r = N
\rightarrow S \slash E'$, where $S$ is the graph
$$
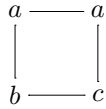
\xymatrix @=20pt{
a \ar@{-}[r] & a \ar@{-}[d] \\
b \ar@{-}[u] & c \ar@{-}[l]
}
$$
and $E' = \{(a,b), (b,a), (c,c), (a,N), (c,N) \}$. (Note that
formally we have only defined $S$ up to isomorphism, however as we
have seen this is not an objection.) Then
Figure~\ref{ex_fig_touching} depicts one possible derivation from
$G$ to a graph $G'$ (on the right-hand side of the figure) for which
no rule is applicable anymore. Note that there is one other possible
derivation to a ``terminal'' graph $G''$ (i.e., a graph without
vertices labelled by $N$): to obtain $G''$ we choose first the
right-hand vertex labelled by $N$ (the one not connected to the
vertex labelled by $b$) in $G$ in the derivation. Note that $G'$ and
$G''$ are different graphs. We assume that the set of labels $L$ is
$\{a,b,c,N\}$. This example will be our running example of this
paper.
\end{Example}

In \cite{Blockeel_NLC_Induction} the inference of NLC grammars with
exactly one rule $r = N \rightarrow S / E$ are studied where
moreover $S$ does not contain a vertex labelled by $N$ and $E$ does
not contain a tuple containing $N$. This is sufficient for the case
where the subgraphs isomorphic to $S$ are non-touching. To consider
the case where the subgraphs are disjoint, we allow $E$ to contain
tuples containing $N$. However, we do require that $S$ remains
without vertices labelled by $N$. Therefore there is no ``real''
recursion: no vertices labelled by $N$ can be introduced in any
derivation.

\section{Known results: Non-touching graphs}
\label{sec_non_touching} In this section we recall some notions and
a result from \cite{Blockeel_NLC_Induction} which we will need in
subsequent sections. First we define in this context the notion of
compatibility.

\begin{Definition}
Let $G$ be a graph and $S$ be an induced subgraph of $G$. We say
that $E \subseteq L \times L$ is \emph{compatible} for $S$ (in $G$)
if there is a graph $F$ such that an application of NLC grammar rule
$N \rightarrow S' \slash E$ to $F$ ``creates'' $S$ and obtains graph
$G$. Note: $S'$ is (isomorphic to) $S$.
\end{Definition}

\begin{Example} \label{ex_nlc_grammar2}
Reconsider our running example. Hence we again let $G'$ be the graph
at the right-hand side of Figure~\ref{ex_fig_touching}. Moreover we
let $S_1$ and $S_2$ be the subgraphs of $G'$ of the form
$$ 
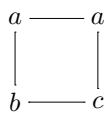
\xymatrix @=20pt{
a \ar@{-}[r] & a \ar@{-}[d] \\
b \ar@{-}[u] & c \ar@{-}[l]
}
$$
where $S_1$ is the one connected to the a vertex labelled by $b$ and
$S_2$ is the other one. Note that $S_1$ and $S_2$ are disjoint and
touching in $G'$. We have that, e.g., $E_1 = \{(b,a), (c,c)\}$, $E_2
= \{(b,a), (c,c), (a,b)\}$ or $E_3 = \{(b,a), (c,c), (c,b), (b,b)\}$
is compatible for $S_2$ in $G'$. The middle graph of the figure is a
graph $F$ such that an application of the NLC grammar rule $N
\rightarrow S \slash E$ to $F$ ``creates'' $S_2$ and obtains graph
$G'$.
\end{Example}

To characterize the notion of compatibility, the notions of
\emph{inset} and \emph{outset} for arbitrary $Q \subseteq V^2$
(where $V$ is the set of edges of $G$) are crucial.
\begin{Definition} \label{inoutset_general}
Let $Q \subseteq V^2$, and let $P_Q = \{In_Q,Out_Q\}$ be the
partition of $Q$ where, for $x \in Q$, $x \in In_Q$ iff $u(x) \in
E(G)$. We define the elements of $\{l(In_Q),l(Out_Q)\}$ as the
\emph{inset}, denoted by $I_Q$, and \emph{outset}, denoted by $O_Q$,
of $Q$, respectively.
\end{Definition}
Let $S$ be an induced subgraph of $G$. Then the \emph{inset}
(\emph{outset}, resp.) of $S$, denoted by $I_S$ ($O_S$, resp.), is
defined to be the inset (outset, resp.) of $Q = K_{V(S),N_G(V(S))}$.

The following lemma, given and proven in
\cite{Blockeel_NLC_Induction}, characterizes compatibility for a
single graph $S$ in terms of the inset and outset of $S$: the inset
are tuples that \emph{should} be in $E$, while the outset are tuples
that \emph{should not} be in $E$.
\begin{Lemma} \label{single_compatible}
Let $S$ be an induced subgraph of $G$, and let $E \subseteq L \times
L$. Then $E$ is compatible for $S$ iff $I_S \subseteq E \subseteq
L^2 \backslash O_S$ (i.e., $E$ separates $I_S$ from $O_S$).
\end{Lemma}
Hence, there is a compatible $E$ for $S$ in $G$ iff $I_S \cap O_S =
\emptyset$.

\begin{Example} \label{ex_nlc_grammar3}
Reconsider again our running example. Then $I_{S_2} = \{(b,a),
(c,c)\}$ and $O_{S_2} = \{(a,a), (a,c), (c,a), (b,c)\}$ (w.r.t.
$G'$). Since $I_{S_2} \cap O_{S_2} = \emptyset$, there is a
compatible $E$ for $S_2$ in $G'$. We have that $I_{S_2} \subseteq E
\subseteq L^2 \backslash O_{S_2}$ holds for, e.g., $E_1$, $E_2$ and
$E_3$ in Example~\ref{ex_nlc_grammar2}.
\end{Example}

We consider now sequences of subgraphs to be generated by a single
graph rule. Note that these graphs must necessarily be mutually
isomorphic.
\begin{Definition}
Let $G$ be a graph and $S_1, S_2, \ldots, S_n$ be induced subgraphs
of $G$ isomorphic to $S$. We say that $E \subseteq L \times L$ is
\emph{compatible} for $(S_1, S_2, \ldots, S_n)$ (in $G$) if there
are graphs $G_0, \ldots, G_n$ such that $G_n = G$ and for each $i
\in \{1,\ldots,n\}$, $G_{i}$ is obtained from $G_{i-1}$ by applying
NLC grammar rule $N \rightarrow S \slash E$ that ``creates'' $S_i$.
\end{Definition}
Note that, in general, the order of the elements $(S_1, S_2, \ldots,
S_n)$ is important. E.g. a given $E$ may be compatible for
$(S_1,S_2)$ while it is incompatible for $(S_2,S_1)$ (we will see
such an example in the next section).

However, for a set of mutually non-touching and isomorphic subgraphs
$S_i$ for $i \in \{1,\ldots,n\}$ of $G$, the order of the elements
is not important. Thus, $E \subseteq L \times L$ compatible for $C =
(S_1, S_2, \ldots, S_n)$ implies that $E$ is compatible for any
permutation of $C$. In fact we have that, $E \subseteq L \times L$
is compatible for $S_1$, for $S_2$, $\ldots$, and for $S_n$ iff it
is compatible for $C$ (or any permutation of $C$). Therefore, in
this case, Lemma~\ref{single_compatible} is trivially generalized:
$E \subseteq L \times L$ is compatible for $(S_1, S_2, \ldots, S_n)$
iff $\cup_i I_{S_i} \subseteq E \subseteq L^2 \backslash (\cup_i
O_{S_i})$ (as noted in \cite{Blockeel_NLC_Induction}).

\section{Two touching graphs} \label{sec_touching_two}
In this section we consider the case where a single NLC grammar rule
$N \rightarrow S \slash E$ generates disjoint subgraphs which can
(possibly) touch each other. Hence, this generalizes
Lemma~\ref{single_compatible} by replacing the non-touching
condition into disjointness. To this aim we allow non-terminal $N$
to be present in tuples of the embedding relation $E$ of NLC grammar
rule $N \rightarrow S \slash E$. This introduces the issue of
confluency: the order in which non-terminals are replaced by
subgraphs influences the obtained graph.
Example~\ref{ex_nlc_grammar} illustrates this as the different
graphs $G'$ and $G''$ can both be obtained from the original graph
$G$.

As we will see the inset and outset between the vertices of two
touching graphs turns out to be crucial.
\begin{Definition} \label{def_inoutset_touching}
Let $S_1$ and $S_2$ be touching graphs in $G$. For $Q_1 =
K_{V(S_2),V(S_1)\cap N_G(S_2)}$, we denote $I_{Q_1}$ and $O_{Q_1}$
by $I_{\spair}$ and $O_{\spair}$, respectively. Moreover, for $Q_2 =
K_{V(S_2),V(S_1)}$, we denote $I_{Q_2}$ and $O_{Q_2}$ by
$I_{(\spair)}$ and $O_{(\spair)}$, respectively.
\end{Definition}
We now state some basic properties of the insets and outsets of
Definition~\ref{def_inoutset_touching}. Note first that $I_{\spair}
= I_{(\spair)}$. In fact, it is equal to the inset of
$$
K_{V(S_2) \cap N_G(S_1),V(S_1) \cap N_G(S_2)}.
$$
Also note that, for node labels $x$ and $y$, we have $(x,y) \in
I_{(\spair)}$ iff $(y,x) \in I_{((S_2,S_1))}$. This holds similarly
for $O_{(\spair)}$, however, this does not hold in general for
$O_{\spair}$. Moreover note that $O_{\spair} \subseteq
O_{(\spair)}$, and
$$
O_{(\spair)} \backslash O_{\spair} = l(K_{V(S_2),V(S_1) \backslash
N_G(S_2)}).
$$
Finally note that $\pi_2(I_{\spair}) = l(V(S_1) \cap N_G(S_2))$.
We will use these basic properties \remconf{frequently} in the
remainder of this paper.

\begin{Example}
In our running example, we have $I_{\spair} = \{(b,a), (c,c)\}$,
$O_{\spair} = \{(a,a), (a,c), (b,c), (c,a)\}$, and $O_{(\spair)} =
L'^2 \backslash I_{\spair}$ with $L' = \{a,b,c\}$. Moreover, we have
$I_{\spairinv} = \{(a,b), (c,c)\}$, $O_{\spairinv} = \{(a,c), (b,b),
(b,c), (c,b)\}$, and $O_{(\spairinv)} = L'^2 \backslash
I_{\spairinv}$.
\end{Example}

We now adapt the definition of inset and outset for a graph $S$, by
incorporating the issues related to touching graphs.
\begin{Definition} \label{def_inoutset_nontouching}
Let $S_1,\dots,S_n$ be distinct subgraphs of $G$, and let $Q =
\cup_{i \in \{1,\ldots,n\}}$ $K_{V(S_i),N_G(V(S_i))\backslash
(\cup_{j \in \{1,\ldots,n\}} V(S_j))}$. We denote $I_Q$ and $O_Q$ by
$I_{[S_1,\ldots,S_n]}$ and $O_{[S_1,\ldots,S_n]}$, respectively.
\end{Definition}

Note that $I_{[S_1,S_2]} = I_{[S_2,S_1]}$ and if $S_1$ and $S_2$ are
non-touching, we have $I_{[S_1,S_2]} = I_{S_1} \cup I_{S_2}$.

\begin{Example}
In our running example, we have $I_{[S_1,S_2]} = \{(a,b)\}$, and
$O_{[S_1,S_2]} = \{(b,b),(c,b)\}$.
\end{Example}

Definitions~\ref{def_inoutset_touching} and
\ref{def_inoutset_nontouching} are to separate three types of insets
and outsets. Roughly speaking, the two types of insets and outsets
of Definition~\ref{def_inoutset_touching} deal with the tuples
between $S_1$ and $S_2$, while the type of inset and outset of
Definition~\ref{def_inoutset_nontouching} deals with the tuples from
$S_1$ to the ``outside world'' (the vertices in the neighborhood of
$S_1$ which do not belong to $S_2$) \emph{plus} the tuples from
$S_2$ to the ``outside world'' (the vertices in the neighborhood of
$S_2$ which do not belong to $S_1$).

We now characterize the embedding relations $E$ such that $E$ is
compatible for $(S_1,S_2)$ where $S_1$ and $S_2$ are touching
subgraphs of $G$.
\begin{Lemma} \label{touching_compatible}
Let $S_1$ and $S_2$ be touching subgraphs of $G$. Then $E \subseteq
L \times L$ is compatible for $(S_1,S_2)$ iff the following
conditions hold:
\begin{enumerate}
\item $I_{\spair} \subseteq E$,

\item $\{(x,N) \mid x \in \pi_2(I_{\spair})\} \subseteq E$,

\item If $e \in O_{(\spair)}$, then either $(\pi_2(e),N) \not\in E$ or $e
\not\in E$ (or both), and

\item
$I_{[S_1,S_2]} \subseteq E \subseteq L^2 \backslash
(O_{[S_1,S_2]})$.
\end{enumerate}
Moreover, if this is the case, then we have $E \cap O_{\spair} =
\emptyset$.
\end{Lemma}
\remconf{
\begin{Proof}
In the case where there are no edges between $S_1$ and $S_2$, we
have, by Lemma~\ref{single_compatible}, that $E \subseteq L \times
L$ is compatible for $(S_1,S_2)$ iff $I_{S_1} \cup I_{S_2} \subseteq
E \subseteq L^2 \backslash (O_{S_1} \cup O_{S_2})$ -- this is
equivalent to condition \emph{(4)} (since $S_1$ and $S_2$ are
non-touching).

Now, since edges between $S_1$ and $S_2$ can only introduce
\emph{additional} constraints on $E$ (i.e., not less constraints),
we may consider only the graph $F = N-N$, an edge having two
vertices labelled by $N$, and check the necessary and sufficient
(additional) constraints on $E$ to transform the graph in two steps
where $S_1$ appears first and then $S_2$ such that the edges between
$S_1$ and $S_2$ are identical to those between $S_1$ and $S_2$ in
$G$.

Now, let $x$ be a vertex of $S_1$ labelled by $b$, and $y$ be a
vertex of $S_2$ labelled by $a$. Assume first that $x$ is connected
to $y$ in $G$. Now, if we apply the NLC rule to create $S_1$, then
$x$ should be connected to $N$ -- thus we need $(b,N) \in E$.
Indeed, without this rule $x$ will not be connected to any vertex of
$S_2$ (after applying the NLC rule to create $S_2$). Now, if we
subsequently apply the NLC rule to create $S_2$, then $y$ should be
connected to $x$ and hence we need $(a,b) \in E$. Hence $(a,b) \in
E$ and $(b,N) \in E$ results in an edge between $x$ and $y$.
Conversely, if either $(b,N) \not\in E$ or $(a,b) \not\in E$, then
$x$ is not connected to $y$. Consequently, both $(a,b) \in E$ and
$(b,N) \in E$ iff there is an edge between a/every vertex labelled
by $b$ in $S_1$ and a/every vertex labelled by $a$ in $S_2$.

Thus, $E$ is compatible for $(S_1,S_2)$ iff $I_{(\spair)} \cup
\{(x,N) \mid x \in \pi_2(I_{(\spair)})\} \subseteq E$ \emph{ and }
both $(a,b) \in E$ and $(b,N) \in E$ implies $(a,b) \not\in
O_{(\spair)}$.

Finally, we have in this case $E \cap O_{\spair} = \emptyset$.
Indeed, if $e \in E \cap O_{\spair}$, then $e \in O_{(\spair)}$ and
$e \in E$ and therefore, by condition \emph{(3)}, $(\pi_2(e),N)
\not\in E$. Now, $\pi_2(O_{\spair}) \subseteq \pi_2(I_{\spair}) =
l(V(S_1) \cap N_G(S_2))$ cf. Definition~\ref{def_inoutset_touching}.
Consequently, by condition \emph{(2)}, $(\pi_2(e),N) \in E$ -- a
contradiction.
\end{Proof}
}

Intuitively, condition \emph{(4)} of Lemma~\ref{touching_compatible}
deals with the edges of $S_1$ and $S_2$ to the ``outside world'',
while conditions \emph{(1)} to \emph{(3)} deal with the edges
between $S_1$ and $S_2$. Conditions \emph{(1)} and \emph{(2)} state
the tuples that must necessarily be in $E$, while condition
\emph{(3)} states requirements on which tuples must not (together)
be in $E$.

Since $E \cap O_{\spair} = \emptyset$ by
Lemma~\ref{touching_compatible}, we may modify conditions \emph{(1)}
and \emph{(3)} of the previous lemma as follows:

\begin{enumerate}
\item[\emph{1'.}]
$I_{\spair} \subseteq E \subseteq L^2 \backslash O_{\spair}$,

\item[\emph{3'.}]
If $e \in O_{(\spair)} \backslash O_{\spair} = l(K_{V(S_2),V(S_1)
\backslash N_G(S_2)})$, then either $(\pi_2(e),N) \not\in E$ or $e
\not\in E$ (or both).
\end{enumerate}
However, in this way the condition $E \cap O_{\spair} = \emptyset$
is explicitly assumed and not part of the result as stated in the
lemma.

\remconf{
\begin{Remark}
By condition \emph{(4)} of the lemma, we may go even further and
instead state ``$e \in l(K_{V(S_2),V(S_1) \backslash N_G(S_2)})
\backslash O_{[S_1,S_2]}$'' in condition \emph{(3)}. Therefore in
practise may be easier to check condition \emph{(3)} if one
considers only the (smaller) set $O_{(\spair)} \backslash
(O_{\spair} \cup O_{[S_1,S_2]}) = l(K_{V(S_2),V(S_1) \backslash
N_G(S_2)}) \backslash O_{[S_1,S_2]}$.
\end{Remark}
}

\remconf{Also note that, we have, for $e \in I_{[S_1,S_2]} \cup
I_{\spair}$ (and hence $e \subseteq E$), $e \in O_{(\spair)}$
implies $(\pi_2(e),N) \not\in E$.}



\begin{Example} \label{ex_touching}
We continue our running example. As we have seen, an $E \subseteq L
\times L$ compatible for $(S_1,S_2)$ in $G'$ allows, given the graph
$G$ on the left-hand side of Figure~\ref{ex_fig_touching}, for the
generation of the middle graph (in the figure) and subsequently the
generation of $G'$. We will now determine, using
Lemma~\ref{touching_compatible} and the modified conditions below
the lemma, the constraints on $E$ for it to be compatible for
$(S_1,S_2)$.

Recall that $I_{\spair} = \{(b,a), (c,c)\}$, $O_{\spair} = \{(a,a),
(a,c), (b,c), (c,a)\}$, $I_{[S_1,S_2]} = \{(a,b)\}$, and
$O_{[S_1,S_2]} = \{(b,b), (c,b)\}$. Moreover, $\{(x,N) \mid x \in
\pi_2(I_{\spair})$ $=$ $l(V(S_1) \cap N_G(S_2))\} =
\{(a,N),(c,N)\}$. Hence, by conditions \emph{(1')}, \emph{(2)}, and
\emph{(4)} of Lemma~\ref{touching_compatible} we have
$$
\{(a,b), (b,a), (c,c), (a,N), (c,N)\} \subseteq E
$$
and
$$
E \cap \{(a,a), (a,c), (b,b), (b,c), (c,a), (c,b)\} = \emptyset.
$$
Now, $O_{(\spair)} \backslash O_{\spair} = l(K_{V(S_2),V(S_1)
\backslash N_G(S_2)}) = l(K_{\{a,b,c\},\{b\}})$ $=$ $\{(a,b),$
$(b,b),$ $(c,b)\}$.
%
%
Hence by condition \emph{(3')} either $(b,N) \not\in E$ or $(a,b)
\not\in E$. The latter is a contradiction, hence $(b,N) \not\in E$.
Consequently,
$$
E = \{(a,b), (b,a), (c,c), (a,N), (c,N)\}
$$
is compatible for $(S_1,S_2)$, in fact, in this case, it is the
unique $E$ such that it is compatible for $(S_1,S_2)$ in $G'$. Note
that adding $(b,N)$ to $E$ would indeed make it incompatible -- the
generated graph would then have edges from the vertex labelled $b$
in $S_1$ to the two vertices labelled $a$ in $S_2$. Also note that
this $E$ is \emph{not} compatible for $(S_2,S_1)$ in $G'$.
\end{Example}

Using Lemma~\ref{touching_compatible}, the \emph{existence} of an
embedding relation $E$ is elegantly characterized, as shown in the
next lemma.

\begin{Lemma} \label{touching_existence}
Let $S_1$ and $S_2$ be touching graphs. There is a compatible $E
\subseteq L \times L$ for $(S_1,S_2)$ iff $(I_{[S_1,S_2]} \cup
I_{\spair}) \cap O_{[S_1,S_2]} = \emptyset$, $\pi_2(I_{\spair}) \cap
\pi_2(I_{[S_1,S_2]} \cap O_{(\spair)}) = \emptyset$, and $I_{\spair}
\cap O_{(\spair)} = \emptyset$. Moreover, if this is the case, then
$(I_{[S_1,S_2]} \cup I_{\spair}) \cap O_{\spair} = \emptyset$.
\end{Lemma}
\remconf{
\begin{Proof}
Assume first that that right-hand side holds. Then take $E' =
I_{[S_1,S_2]} \cup I_{\spair}$, take $F' = \pi_2(I_{\spair})$, and
let $E = E' \cup \{(x,N) \mid x \in F'\}$. Now, conditions
\emph{(1)}, \emph{(2)}, and \emph{(4)} of
Lemma~\ref{touching_compatible} hold trivially. Finally to prove
condition \emph{(3)}, we need to show that $e \in O_{(\spair)} \cap
E$ implies $(\pi_2(e),N) \not\in E$. Let $e \in O_{(\spair)} \cap
E$. We have, by definition of $E$, that $e \in I_{[S_1,S_2]}$ or $e
\in I_{\spair}$. The latter is a contradiction of $I_{\spair} \cap
O_{(\spair)} = \emptyset$. The former implies, by the second
equation of this lemma, that $\pi_2(e) \not\in \pi_2(I_{\spair}) =
F'$. Consequently, $(\pi_2(e),N) \not\in E$.

Now, we prove the other implication. If there is such compatible
$E$, then, by Lemma~\ref{touching_compatible}, $(I_{[S_1,S_2]} \cup
I_{\spair}) \cap O_{[S_1,S_2]} = \emptyset$. Assume $I_{\spair} \cap
O_{(\spair)} \not= \emptyset$, and let $e \in I_{\spair} \cap
O_{(\spair)}$. Since $e \in I_{\spair}$, we have, by condition
\emph{(1)} in Lemma~\ref{touching_compatible}, $e \in E$, and we
have by condition \emph{(2)} $(\pi_2(e),N) \in E$. Now since $e \in
O_{(\spair)}$ we have a contradiction by condition \emph{(3)}.
Finally, assume $\pi_2(I_{\spair}) \cap \pi_2(I_{[S_1,S_2]} \cap
O_{(\spair)}) \not= \emptyset$ and let $x \in \pi_2(I_{\spair}) \cap
\pi_2(I_{[S_1,S_2]} \cap O_{(\spair)})$. Then, by condition
\emph{(2)}, $(x,N) \in E$, and by condition \emph{(3)}, $(x,N)
\not\in E$ -- a contradiction.

By Lemma~\ref{touching_compatible}, we have in this case
$(I_{[S_1,S_2]} \cup I_{\spair}) \cap O_{\spair} = \emptyset$, since
$I_{[S_1,S_2]} \cup I_{\spair} \subseteq E$ and $E \cap O_{\spair} =
\emptyset$.
\end{Proof}
}

Recall that $I_{\spair} = I_{(\spair)}$, hence the third equation of
Lemma~\ref{touching_existence} may be rephrased more symmetrically
as ``$I_{(\spair)} \cap O_{(\spair)} = \emptyset$''. Notice that the
case $N_G(V(S_1) \cup V(S_2)) = \emptyset$ (roughly) corresponds to
the situation where the original graph $F$ that generates $G$ has a
connected component $N-N$. In this case, by
Lemma~\ref{touching_existence}, there is a compatible $E \subseteq L
\times L$ for $(S_1,S_2)$ iff $I_{(\spair)} \cap O_{(\spair)} =
\emptyset$ (since $I_{[S_1,S_2]} = O_{[S_1,S_2]} = \emptyset$).

\begin{Example}
We continue Example~\ref{ex_touching} (our running example). Recall
that $I_{[S_1,S_2]} \cup I_{\spair} = \{ (a,b), (b,a), (c,c) \}$ and
$O_{[S_1,S_2]} = \{(b,b), (c,b)\}$ -- hence they are disjoint. Also,
$\pi_2(I_{\spair}) = \{a,c\}$ and $\pi_2(I_{[S_1,S_2]} \cap
O_{(\spair)}) = \pi_2(\{(a,b)\}) = \{b\}$, and therefore they are
disjoint. Finally, $I_{\spair} \cap O_{(\spair)} = \emptyset$.
Consequently, by Lemma~\ref{touching_existence}, there is a
compatible $E$ for $(S_1,S_2)$ -- such an $E$ is given in
Example~\ref{ex_touching}.
\end{Example}


\section{Set of touching graphs}
Let $\mathcal{S} = \{S_i \mid i \in \{1,\ldots,n\}\}$ be a set of
mutually isomorphic and disjoint subgraphs of $G$. In this section
we turn to the question of whether or not there is an $E \subseteq L
\times L$ \emph{and} a linear ordering $C = (S_{i_1}, S_{i_2},
\ldots, S_{i_n})$ of $\mathcal{S}$ such that $E$ is a compatible
embedding relation for $C$.

The following result is easily obtained from
Lemma~\ref{sec_touching_two}.
\begin{Lemma} \label{lem_seq_graphs_char_E}
Let $G$ be a graph, $E \subseteq L \times L$, and $C =
(S_1,\ldots,S_n)$ be a sequence of mutually disjoint induced
subgraphs of $G$ isomorphic to $S$. Then $E$ is compatible for $C$
iff (1) $I_{[S_1,\ldots,S_n]} \subseteq E \subseteq L^2 \backslash
(O_{[S_1,\ldots,S_n]})$ and (2) for each two touching $S_i$ and
$S_j$ with $i<j$, we have that the first three conditions of
Lemma~\ref{touching_compatible} hold w.r.t. $E$ and $(S_i,S_j)$.
\end{Lemma}

Clearly, if $S_i$ and $S_{i+1}$ are non-touching, then $E$ is
compatible for $(S_1, \ldots,$ $S_i,$ $S_{i+1},$ $\ldots, S_n)$ iff
$E$ is compatible for $(S_1, \ldots, S_{i+1}, S_i, \ldots, S_n)$.
Thus, as we have already seen in Section~\ref{sec_non_touching}, the
case where $S_1, S_2, \ldots, S_n$ are mutually non-touching is much
less involved: $E$ is compatible for each linear ordering of
$\mathcal{S}$. For touching graphs, the situation is different as
the conditions in Lemma~\ref{touching_compatible} are not symmetric:
e.g. $I_{(S_i,S_j)}$ and $I_{(S_j,S_i)}$ generally differ. Hence, we
must choose a linear ordering in a ``compatible'' way. First, we
focus on the question whether or not there exists an $E$ compatible
for a given linear ordering $C$ of $\mathcal{S}$. We characterize
the existence by generalizing Lemma~\ref{touching_existence} for the
case where more than two graphs can touch each other. To this aim
consider the following graph that represents whether or not
subgraphs $S_i$ and $S_j$ in $\mathcal{S}$ touch.

\begin{Definition}
Let $G$ be a graph and $\mathcal{S} = \{S_i \mid i \in
\{1,\ldots,n\}\}$ be a set of induced subgraphs of $G$. The
\emph{touching graph} of $G$ w.r.t. $\mathcal{S}$, is the
(undirected) graph $(\mathcal{S},\{\{S_i,S_j\} \mid S_i \mbox{ and }
S_j \mbox{ touch} \})$.
\end{Definition}

We now give the edges of a touching graph an orientation such that
the obtained graph, called directed touching graph, is acyclic.
\begin{Definition}
Let $T$ be the touching graph of $G$ w.r.t. $\mathcal{S}$. Then the
\emph{directed touching graph} of $G$ w.r.t. to an ordering
$(S_1,S_2,\ldots,S_n)$ of $\mathcal{S}$ is the directed graph $D =
(V(D),E(D))$ where, $V(D) = V(T)$ and $(S_i,S_j) \in E(D)$ iff
$\{S_i,S_j\} \in E(T)$ and $i < j$.
\end{Definition}

For $e = (S_i,S_j) \in E(D)$ we write $O_e = O_{(S_i,S_j)}$ and
$O_{(e)} = O_{((S_i,S_j))}$ (and similarly for the insets $I_e$ and
$I_{(e)}$).

We now obtain the main result -- it generalizes
Lemma~\ref{touching_existence}.
\begin{Theorem} \label{th_main_char_existence}
Let $G$ be a graph and $C = (S_1,\ldots,S_n)$ be a sequence of
induced subgraphs of $G$ and let $D$ be the directed touching graph
of $G$ w.r.t. $C$. There is a compatible $E \subseteq L \times L$
for $C$ iff
\begin{eqnarray}
(I_{[S_1,\ldots,S_n]} \cup (\cup_{e\in E(D)} I_{e})) \cap
O_{[S_1,\ldots,S_n]} = \emptyset,
\\
\pi_2(\cup_{e\in E(D)} I_{e}) \cap \pi_2(I_{[S_1,\ldots,S_n]} \cap
(\cup_{e\in E(D)} O_{(e)})) = \emptyset, \mbox{ and}
\\
(\cup_{e\in E(D)} I_{e}) \cap (\cup_{e\in E(D)} O_{(e)}) =
\emptyset.
\end{eqnarray}
Moreover, if this is the case, then $(I_{[S_1,\ldots,S_n]} \cup
(\cup_{e\in E(D)} I_{e})) \cap (\cup_{e\in E(D)} O_{e}) =
\emptyset$.
\end{Theorem}
\remconf{
\begin{Proof}
This proof will be in the same spirit as the proof of
Lemma~\ref{touching_existence}.

Assume first that that right-hand side holds. Then take $E' =
I_{[S_1,\ldots,S_n]} \cup (\cup_{e\in E(D)} I_{e})$, and take $F' =
\pi_2(\cup_{e\in E(D)} I_{e})$. Now, let $E = E' \cup \{(x,N) \mid x
\in F'\}$. By Lemma~\ref{lem_seq_graphs_char_E} it suffices to show
that for each two touching $S_i$ and $S_j$ with $i<j$, the first
three conditions of Lemma~\ref{touching_compatible} hold w.r.t. $E$
and $r = (S_i,S_j)$. Now, conditions \emph{(1)}, \emph{(2)}, and
\emph{(4)} of Lemma~\ref{touching_compatible} hold trivially.
Finally to prove condition \emph{(3)}, we need to show that $e \in
O_{(r)} \cap E$ implies $(\pi_2(e),N) \not\in E$. Let $e \in O_{(r)}
\cap E$. We have, by definition of $E$, that $e \in
I_{[S_{k_1},S_{k_2}]}$ or $e \in I_{(S_{k_3},S_{k_4})}$ for some
$k_1, \ldots, k_4$. The latter is a contradiction of
$I_{(S_{k_3},S_{k_4})} \cap O_{(r)} = \emptyset$. The former implies
by the second equation of this theorem that $\pi_2(e) \not\in
\pi_2(\cup_{e\in E(D)} I_{e}) = F'$. Consequently, $(\pi_2(e),N)
\not\in E$.

Now, we prove the other implication. Assume that there is a
compatible $E \subseteq L \times L$ for $C$. Then by
Lemma~\ref{lem_seq_graphs_char_E}, (1) $I_{[S_1,\ldots,S_n]}
\subseteq E \subseteq L^2 \backslash (O_{[S_1,\ldots,S_n]})$ and (2)
for each two touching $S_i$ and $S_j$ with $i<j$, we have that the
first three conditions of Lemma~\ref{touching_compatible} hold
w.r.t. $E$ and $(S_i,S_j)$. Hence, by
Lemma~\ref{touching_compatible}, $(I_{[S_1,\ldots,S_n]} \cup
(\cup_{e\in E(D)} I_{e})) \cap O_{[S_1,\ldots,S_n]} = \emptyset$.
Assume now that $I_{f_1} \cap O_{(f_2)} \not= \emptyset$ for some
$f_1,f_2 \in E(D)$, and let $e \in I_{f_1} \cap O_{(f_2)}$. Since $e
\in I_{f_1}$, we have, by condition \emph{(1)} in
Lemma~\ref{touching_compatible}, $e \in E$, and we have by condition
\emph{(2)} $(\pi_2(e),N) \in E$. Now since $e \in O_{(f_2)}$ we have
a contradiction by condition \emph{(3)}. Finally, assume
$\pi_2(I_{f_1}) \cap \pi_2(I_{[S_1,\ldots,S_n]} \cap O_{(f_2)})
\not= \emptyset$ and let $x \in \pi_2(I_{f_1}) \cap
\pi_2(I_{[S_1,\ldots,S_n]} \cap O_{(f_2)})$. Then, by condition
\emph{(2)}, $(x,N) \in E$, and by condition \emph{(3)}, $(x,N)
\not\in E$ -- a contradiction.

Finally, by Lemma~\ref{touching_existence}, if this is the case,
then $(I_{[S_1,\ldots,S_n]} \cup (\cup_{e\in E(D)} I_{e})) \cap
(\cup_{e\in E(D)} O_{e}) = \emptyset$.
\end{Proof}
}

\section{Determining compatible sequences of subgraphs}
In this section we turn to the question of efficiently determining,
given a set $\mathcal{S} = \{S_i \mid i \in \{1,\ldots,n\}\}$ of
disjoint subgraphs, an ordering $C$ of $\mathcal{S}$ (if it exists)
such that there is a compatible $E \subseteq L \times L$ for $C$.

We proceed as follows. First, assuming such ordering $C$ exists, by
Theorem~\ref{th_main_char_existence}, the following equality
\begin{eqnarray}\label{separation_eqn}
(I_{[S_1,\ldots,S_n]} \cup (\cup_{e\in E(D)} I_{e})) \cap
(O_{[S_1,\ldots,S_n]} \cup (\cup_{e\in E(D)} O_{e})) = \emptyset
\end{eqnarray}
holds, where $D$ is the directed touching graph w.r.t. $C$. We
consider this equality instead of $(I_{[S_1,\ldots,S_n]} \cup
(\cup_{e\in E(D)} I_{e})) \cap O_{[S_1,\ldots,S_n]} = \emptyset$ for
computational efficiency reasons, as we will see below.

Note that, because of distributivity $(A \cup B) \cap C = (A \cap C)
\cup (B \cap C)$, Equation~(\ref{separation_eqn}) is equal to
\begin{eqnarray*}
\left( \bigcup_{e,f \in E(D)} (I_{e} \cap O_{f}) \right) \cup \left(
\bigcup_{e \in E(D)} ((I_{[S_1,\ldots,S_n]} \cap O_{e}) \cup (I_{e}
\cap O_{[S_1,\ldots,S_n]})) \right) \\
\cup (I_{[S_1,\ldots,S_n]} \cap O_{[S_1,\ldots,S_n]}).
\end{eqnarray*}
Now, for touching graphs $S_1$ and $S_2$, we define $e=(S_1,S_2)$
\emph{admissible} (w.r.t. $\mathcal{S}$) if $(I_{[S_1,\ldots,S_n]}
\cap O_{e}) \cup (I_{e} \cap O_{[S_1,\ldots,S_n]}) = \emptyset$. Or
equivalently,
$$
I_{[S_1,\ldots,S_n]} \cap O_{e} = \emptyset  \mbox{ and } I_{e} \cap
O_{[S_1,\ldots,S_n]} = \emptyset.
$$
Now, to determine the existence of an ordering $C$ of $\mathcal{S}$
and a $E \subseteq L \times L$ such that $E$ is compatible for $C$,
we first check whether or not $I_{[S_1,\ldots,S_n]} \cap
O_{[S_1,\ldots,S_n]} = \emptyset$. If this does not hold, there is
no such $C$ (and $E$). Otherwise, we construct the admissible
touching graph.

\begin{Definition}
Let $T$ be the touching graph of $G$ w.r.t. $\mathcal{S}$. Then the
\emph{admissible touching graph} of $G$ w.r.t. $\mathcal{S}$ is the
directed graph $D = (V(D),E(D))$ where, $V(D) = V(T)$ and, for $e =
(S,S')$ with $S,S' \in V(D)$, $e \in E(D)$ iff $e$ is admissible.
\end{Definition}
Now, since we consider Equation~(\ref{separation_eqn}), this graph
will be considerably smaller than the corresponding graph for the
original equation. This may correspond to a substantial speedup as
we subsequently check conditions between edges of this graph.

Recall that $O_{f} \subseteq O_{(f)}$, and hence $I_{e} \cap O_{(f)}
= \emptyset$ implies $I_{e} \cap O_{f} = \emptyset$. Thus we need to
check for each topological ordering $C$ of the admissible touching
graph whether or not
\begin{eqnarray*}
\bigcup_{e,f \in E(D)} (I_{e} \cap O_{(f)}) = \emptyset, \mbox{ and}
\\
\bigcup_{e,f \in E(D)} (\pi_2(I_{e}) \cap \pi_2(I_{[S_1,\ldots,S_n]}
\cap O_{(f)})) = \emptyset
\end{eqnarray*}
where $D$ is the directed touching graph w.r.t. $C$. If there is
such a $C$, then there is an $E$ compatible for $C$. Otherwise,
there is no linear ordering $C$ and embedding relation $E$ where $E$
is compatible for $C$.

\incomplete{
\section{NLC Graph Grammars with Multiple Productions}
\label{sec_multiple_prod} In this section we consider the general
case where we the NLC graph grammar $Q$ can have more than one
production. Hence we have a set $P$ of productions in the $Q$: $P =
\{A_i \rightarrow S_i \mid i \in \{1,\ldots,n\}\}$. Recall that the
embedding relation $E$ is always fixed for given NLC graph grammar.
We do however again assume that the labels of $S_i$ are
non-terminals, i.e., none of the labels are equal to $A_i$ for any
$i$.

In the special case where all $A_i$ are identical, say to $N$, the
results of the previous sections carry over essentially unchanged.
Indeed, the restriction that all the elements of $\mathcal{S}$ are
isomorphic to some graph $S$ is not relevant in any of the results
and proofs -- the restriction is only needed as we have only one
production $N \rightarrow S$.

However, some modifications are needed to consider the case where we
have different non-terminals $A_i$.

TODO
}

\section{Discussion}
In this paper we considered the problem of graph grammar inference
for the case where one is given a disjoint set $\mathcal{S}$ of
isomorphic subgraphs to be generated by a single rule $r = N
\rightarrow S \slash E$, where the embedding relation $E$ is allowed
to contain tuples containing $N$. In this way we generalize results
in \cite{Blockeel_NLC_Induction}. This result is to be seen as a
further step towards a systematic account of NLC graph grammar
inference.

Formally, we characterized, given a $\mathcal{S} = \{S_i \mid i \in
\{1,\ldots,n\}\}$, the existence of an ordering $C$ of $\mathcal{S}$
and a $E \subseteq L \times L$ such that $E$ is compatible for $C$.
Moreover, if such a $C$ exists, then it is shown to be a topological
ordering of a suitable graph that identifies admissible pairs of
touching subgraphs. The efficiency of the proposed algorithm depends
significantly on the cardinality of $\mathcal{S}$ -- for small
$\mathcal{S}$ the algorithm seems feasible, however this has yet to
be verified in practice.

Finding a graph $S$, such that the set $\mathcal{S}$ of subgraphs of
$G$ isomorphic to $S$ is (1) ``compressible'', i.e. there is a
compatible embedding relation for suitable ordering of
$\mathcal{S}$, and (2) optimal (either in cardinality, or in some
other measure) remains to be investigated.

Also, it is natural to consider the case where for rule $r = N
\rightarrow S \slash E$, $N$ is allowed to be a label on a nodes of
$S$ instead of $N$ contained in (tuples of) $E$. This would have the
consequence that an infinite number of graphs can be generated by
$r$, and, moreover, multiple copies of $S$ can overlap -- loosening
the restriction of disjointness considered here.

\bibliography{../graphtrans}

\end{document}